\documentclass[runningheads]{llncs}
\usepackage[T1]{fontenc}
\usepackage{graphicx}
\usepackage{subfigure}
\usepackage{amsmath} 
\usepackage{amsfonts}
\begin{document}
\title{Synthetic Simplicity: Unveiling Bias in Medical Data Augmentation}
%
%

\author{Krishan Agyakari Raja Babu\inst{1}
 \and
Rachana Sathish\inst{2}\and
Mrunal Pattanaik\inst{2}
\and 
Rahul Venkataramani\inst{2}\email{\{rahul.venkataramani\}@gehealthcare.com}}

\institute{Indian Institute of Technology Madras, India \\ \and
	GE HealthCare, Bangalore, India
	}

\authorrunning{Krishan Agyakari Raja Babu et al.}
%
\maketitle              
\begin{abstract}
Synthetic data is becoming increasingly integral in data-scarce fields such as medical imaging, serving as a substitute for real data. However, its inherent statistical characteristics can significantly impact downstream tasks, potentially compromising deployment performance. In this study, we empirically investigate this issue and uncover a critical phenomenon: downstream neural networks often exploit spurious distinctions between real and synthetic data when there is a strong correlation between the data source and the task label. This exploitation manifests as \textit{simplicity bias}, where models overly rely on superficial features rather than genuine task-related complexities. Through principled experiments, we demonstrate that the source of data (real vs.\ synthetic) can introduce spurious correlating factors leading to poor performance during deployment when the correlation is absent. We first demonstrate this vulnerability on a  digit classification task, where the model spuriously utilizes the source of data instead of the digit to provide an inference. We provide further evidence of this phenomenon in a medical imaging problem related to cardiac view classification in echocardiograms, particularly distinguishing between 2-chamber and 4-chamber views. Given the increasing role of utilizing synthetic datasets, we hope that our experiments serve as effective guidelines for the utilization of synthetic datasets in model training.


\end{abstract}
%
%
\section{Introduction}
\label{sec:intro}
The acute challenge of data scarcity in medical imaging has spurred the development and application of various methods, including transfer learning, self-supervised learning, and data augmentation. More recently, synthetic data, either alone or combined with real data, has emerged as a popular technique to address the inherent data scarcity in medical imaging. The growing prominence of synthetic data is largely attributed to the efficiency of advanced generative techniques, such as generative adversarial networks (GANs), diffusion models ~\cite{Croitoru_2023}, which produce high-quality visual images with relatively straightforward training. As visual generative models continue to improve, researchers are increasingly exploring the utility of generated synthetic samples to augment real data ~\cite{ktena2023generative,stojanovski2023echo}, aiming to achieve fairness and generalization.

Since the images produced by generative models appear extremely realistic, deep learning practitioners have increasingly adopted the use of synthetic data as a substitute of real data. In this paper, we investigate whether synthetic images possess statistical signatures, which may perhaps be inconspicuous visually, specific to their generation algorithms, distinguishing them from those found in real images. These distinctive features could potentially be exploited by downstream algorithms. It is well-known that discriminative deep learning methods tend to exploit the simplest features while disregarding other relevant aspects—a phenomenon referred to as shortcut learning due to ``simplicity bias” ~\cite{shah2020pitfalls}. Medical imaging is particularly prone to these scenarios as data for a particular class could be hard to obtain (e.g.\ pathologies). Since the presence of ``simplicity bias" can lead to sub-optimal performance during deployment, it is necessary for model developers to be cognizant of the perils of indiscriminate use of synthetic data during model development.

In summary, our contributions are as follows:
\begin{itemize}
	\item We conceive the possibility of high correlations between source of data and target resulting in deep learning model exploiting non-task specific features leading to poor performance during deployment.
	\item We expose, with experiments on controlled datasets covering both natural images (digit classification) and medical imaging tasks (echocardiograph view classification), the problem of models exploiting the source of dataset (real vs.\ synthetic) as opposed to the task-specific causal feature.
\end{itemize}

\section{Related Work}
\label{sec:related_work}
Diffusion Probabilistic Models (DDPM), ~\cite{ho2020denoising,sohldickstein2015deep}, have demonstrated remarkable performance in generating high-quality image samples. DDPMs leverage iterative denoising processes to model complex data distributions effectively. The core idea of DDPM revolves around iteratively applying a denoising process to generate samples from a target distribution. To enhance control over semantic details, recent advancements in generative modeling have focused on conditioning models on object-level segmentation masks. A semantic diffusion model (SDM) was proposed in a recent study ~\cite{stojanovski2023echo} for generating synthetic 2D echocardiograms. In this study, we adopt the SDM framework as our model for generating synthetic images.

The promise shown by DDPMs, SDMs and similar models have spurred the utilization of synthetic data  to enrich the training dataset~\cite{fu2024dreamda,stojanovski2023echo}, anomaly detection~\cite{wolleb2022diffusion}, super resolution~\cite{li2022srdiff} etc.
Due to the increasing utilization of synthetic datasets in medical imaging, it is imperative to critically analyze the potential pitfalls carefully. Particularly, the phenomenon of \textit{simplicity bias} has been observed in discriminative deep learning methods, where models tend to exploit spurious, low-level statistical patterns that correlate significantly with the target rather than high-level semantic features, leading to poor out-of-distribution (OOD) performance . This issue is exacerbated in scenarios where synthetic data is used to augment real datasets. Particularly, three implications of the simplicity bias have been discussed by \cite{shah2020pitfalls}—lack of robustness, lack of reliable confidence estimates, and suboptimal generalization.  While the dangers of shortcut learning in the context of healthcare~\cite{geirhos2022imagenettrained,oakdenrayner2019hidden} have been discussed, there is no prior work in the context of spurious features manifested due to utilization of synthetic data.

\section{Methodology}
\label{sec:method}
In order to explore the manifestation of synthetic simplicity in medical data augmentation, we consider the binary classification problem on a dataset $D = \{(\textbf{X}_1, y_1),(\textbf{X}_2, y_2),\dots,(\textbf{X}_N, y_N)\}$ comprising of $N$ samples, where input images $\textbf{X} \in \mathbb{R}^{H \times W}$ and labels $y \in \{-1, +1\}$.Each image can be considered to possess $M$ latent features represented as $f \in \mathbb{R}^M$ which can be leveraged for classification. One of the features can be assumed to be a binary feature $f_k \in \{0, 1\}$, where $k$ denotes the feature index, representing whether a given sample is real or generated synthetically. We intentionally set the correlation between this binary feature and the output label to be high during training and evaluate during the testing phase when the correlation is lower. The test set consists of samples of the form \((\textbf{X}, y = 1, f_k = 0)\), indicating the low correlation between \(y=1\) and \(f_k=0\). We vary the degree of correlation using the parameter $\alpha$ to explore different scenarios of synthetic simplicity.

So, with this configuration, the model during training has the potential to learn the simpler but spurious feature of real vs.\ synthetic. Since the correlation between the task label and data source is quite high, this will result in excellent performance during training. However, during evaluation when the correlation is lower, the performance may degrade if the simplicity bias is manifested. This test allows us to systematically investigate if the neural network utilizes the rest of the features \(f_{i \setminus k}\) or is predisposed to utilize feature \(f_k\) due to it being simpler than other features.

\section{Experiments}
\label{sec:expt}
\subsection{Synthetic Data Engineering}In this section, we will describe the dataset generation for both the digit classification and the echocardiographic view classification problem. The classification network was trained using real data augmented by the synthetic data in varying proportions to study the impact of the synthetic simplicity. Let $D^{Tr}_{R,C_1}$ denote the set of all real images of class 1, $D^{Tr}_{S,C_1}$ denote the set of all synthetic images of class 1, $D^{Tr}_{R,C_2}$ denote the set of all real images of class 2, and $D^{Tr}_{S,C_2}$ denote the set of all synthetic images of class 2. The composition of the training data is thus defined as follows:

\[
|D^{Tr}| = \alpha |D^{Tr}_{R,C_1}| + (1 - \alpha) |D^{Tr}_{S,C_1}| + (1 - \alpha) |D^{Tr}_{R,C_2}| + \alpha |D^{Tr}_{S,C_2}|
\]

where $\alpha \in [0, 1]$ is a hyperparameter and $\alpha|D^{Tr}_{R,C_1}| + (1-\alpha) |D^{Tr}_{S,C_1}| = (1-\alpha)|D^{Tr}_{R,C_2}| + \alpha|D^{Tr}_{S,C_2}|=k_{Tr}$. The test set has an equal number of real and synthetic images for both classes, i.e., $|D^{Ts}_{R,C_1}| = |D^{Ts}_{S,C_1}| = |D^{Ts}_{R,C_2}| = |D^{Ts}_{S,C_2}|=k_{Ts}$.

We design the experiments to include a mix of real and synthetic images of class 1 and 2 in the training dataset $D^{Tr}$, controlled by the hyperparameter $\alpha$.
When $\alpha=1$, the training dataset $D^{Tr}$ consists entirely of real images of class 1 ($D^{Tr}_{R,C_1}$) and synthetic images of class 2 ($D^{Tr}_{S,C_2}$). When \(\alpha = 0\), the training dataset \(D^{Tr}\) consists entirely of synthetic images of class 1 (\(D^{Tr}_{S,C_1}\)) and real images of class 2 (\(D^{Tr}_{R,C_2}\)).

For values of \(\alpha\) between 0 and 1, the training dataset \(D^{Tr}\) is a mix of real and synthetic images of both classes 0 and 1, with \(\alpha\) determining the specific proportion. If the model is able to maintain consistent performance regardless of the value of $\alpha$, it can be considered robust. Otherwise, it may be affected by synthetic simplicity.

\subsubsection{Downstream Network}- In the case of digit classification task, we employed a simple Convolutional Neural Network (CNN)~\cite{oshea2015introduction} as the downstream task. The model was trained on 6000 images each of digits 2 and 4 ($k_{Tr}=6000$) , using the Adam optimizer with a learning rate of 0.001, a batch size of 128, and the cross-entropy loss criterion, over the course of 50 epochs. Evaluation was conducted on a separate dataset comprising 1000 real and synthetic images for each digit 2 and 4 ($k_{Ts}=1000$).

Similarly, for the echocardiogram view classification task, we utilized ResNet-18~\cite{7780459} as the downstream network. The model was trained on 400 views each of the two-chamber and four-chamber echocardiograms ($k_{Tr}=400$) using the Adam optimizer with a learning rate of 0.001, a batch size of 64, and the cross-entropy loss criterion. The training was conducted for 15 epochs. Evaluation was performed on a separate test set containing 100 real and synthetic views each for the two-chamber and four-chamber echocardiograms ($k_{Ts}=100$).

\subsubsection{Diffusion Model}- To generate synthetic digits, we employed the open-source generative diffusion ~\cite{dhariwal2021diffusion,ho2020denoising} model. And, in order to generate synthetic samples for echocardiograms, we utilized a pre-trained modified Semantic Diffusion Model (SDM) provided by ~\cite{stojanovski2023echo}. The synthetic samples were created based on the conditioning of semantic segmentation maps within the SDM framework.

We describe the two datasets used for experiments in the sections below: 

\subsection{MNIST Digit Classification}
\begin{figure}
	\centering
	\subfigure[]{\includegraphics[width=0.24\textwidth]{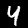}} 
	\subfigure[]{\includegraphics[width=0.24\textwidth]{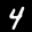}} 
	\subfigure[]{\includegraphics[width=0.24\textwidth]{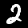}}
	\subfigure[]{\includegraphics[width=0.24\textwidth]{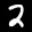}}
	\caption{(a) Digit 4 Real (b) Digit 4 Synthetic (c) Digit 2 Real (d) Digit 2 Synthetic}
	\label{fig:mnist_samples}
\end{figure}

We utilized the publicly available "Modified National Institute of Standards and Technology" (MNIST)~\cite{lecun2010mnist} handwritten digits dataset for our experiment. It consists of 70,000 $28\times28$ black-and-white images of handwritten digits. For our binary classification experiment, we focused on the ``2" vs./ ``4" classes in the MNIST dataset. Figure \ref{fig:mnist_samples} depicts the real and synthetic samples of the digits 2 and 4. 

\subsection{CAMUS Echocardiogram View Classification}

\begin{figure}[]
	\centering
	\subfigure[]{
		\begin{minipage}[b]{0.22\textwidth} 
			\centering
			\includegraphics[width=\textwidth]{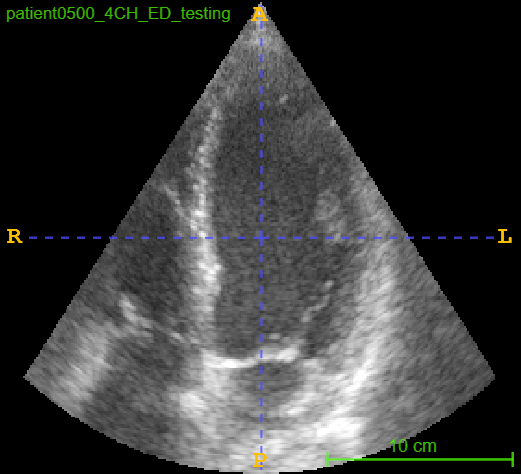}
			
			\includegraphics[width=\textwidth]{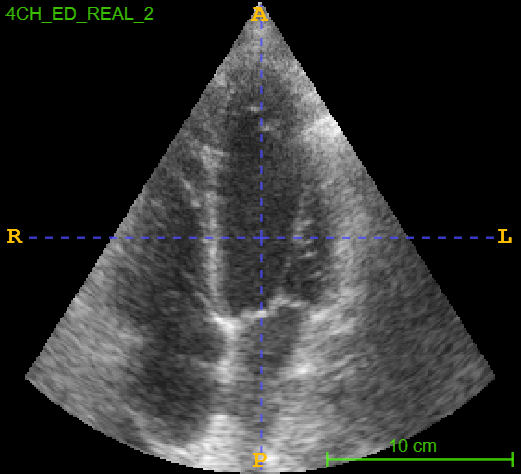}
		\end{minipage}
	}
	\subfigure[]{
		\begin{minipage}[b]{0.22\textwidth}
			\centering
			\includegraphics[width=\textwidth]{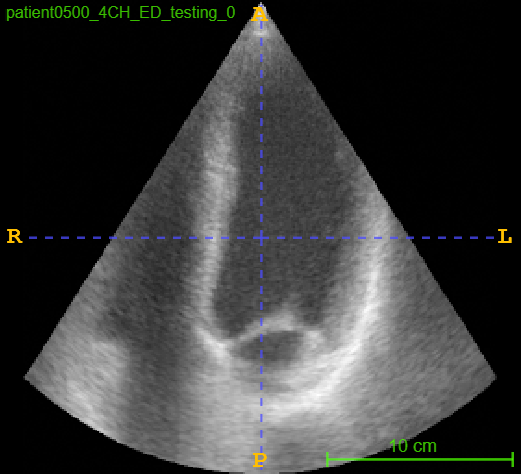}
			
			\includegraphics[width=\textwidth]{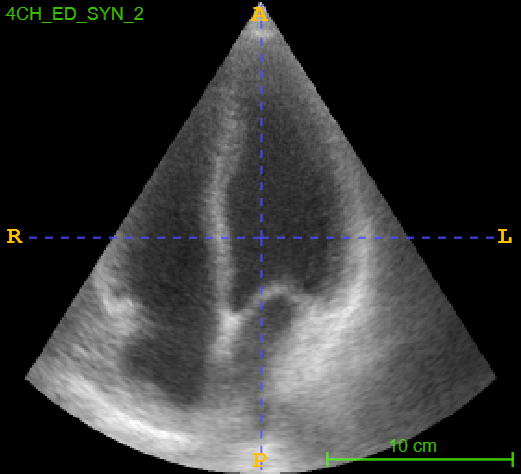}
		\end{minipage}
	}
	\subfigure[]{
		\begin{minipage}[b]{0.22\textwidth} 
			\centering
			\includegraphics[width=\textwidth]{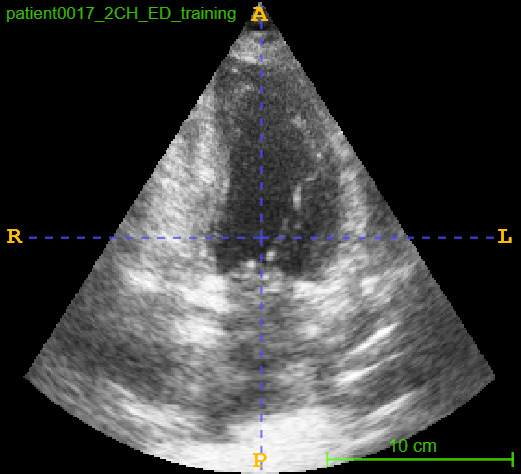}
			
			\includegraphics[width=\textwidth]{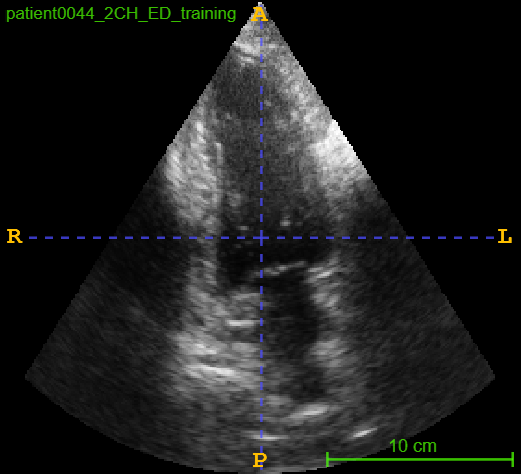}
		\end{minipage}
	}
	\subfigure[]{
		\begin{minipage}[b]{0.22\textwidth}
			\centering
			\includegraphics[width=\textwidth]{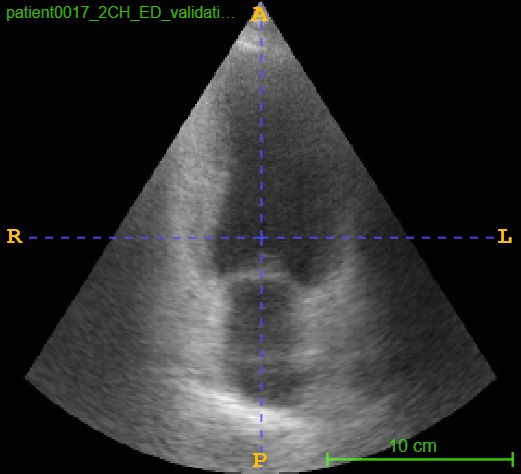}
			
			\includegraphics[width=\textwidth]{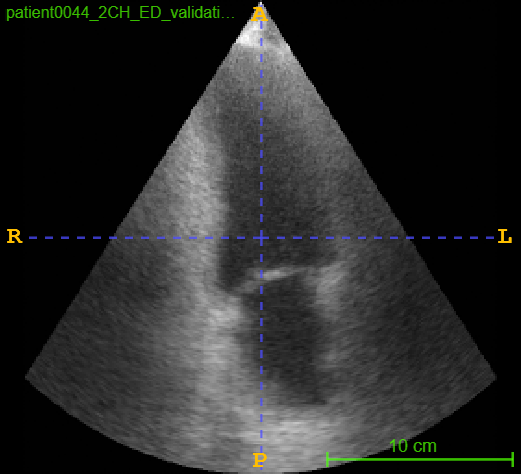}
		\end{minipage}
	}
	\caption{(a) 4CH ED Real, (b) 4CH ED Synthetic, (c) 2CH ED Real, (d) 2CH ED Synthetic}
	\label{fig:camus_samples}
\end{figure}

After investigating the issue of synthetic simplicity in the digit classification task, we demonstrate the phenomenon on a representative medical imaging task of cardiac view classification from echocardiographs. We utilized the publicly available echocardiography dataset "Cardiac Acquisitions for Multi-structure Ultrasound Segmentation" (CAMUS) ~\cite{camus} for this experiment. The CAMUS dataset comprises four types of cardiac views from over 500 subjects: two-chamber end-systolic (2CH ES), two-chamber end-diastolic (2CH ED), four-chamber end-systolic (4CH ES), and four-chamber end-diastolic (4CH ED). We focused on the "4CH ED" and "2CH ED" classes in the CAMUS dataset for classification task. Figure \ref{fig:camus_samples} shows real and synthetic samples of 4CH ED and 2CH ED echocardiogram views. 
\section{Results and Discussion}
\label{sec:results}
\begin{figure}[ht]
	\centering
	
	\subfigure[2CH Real Images]{
		\includegraphics[width=0.45\textwidth]{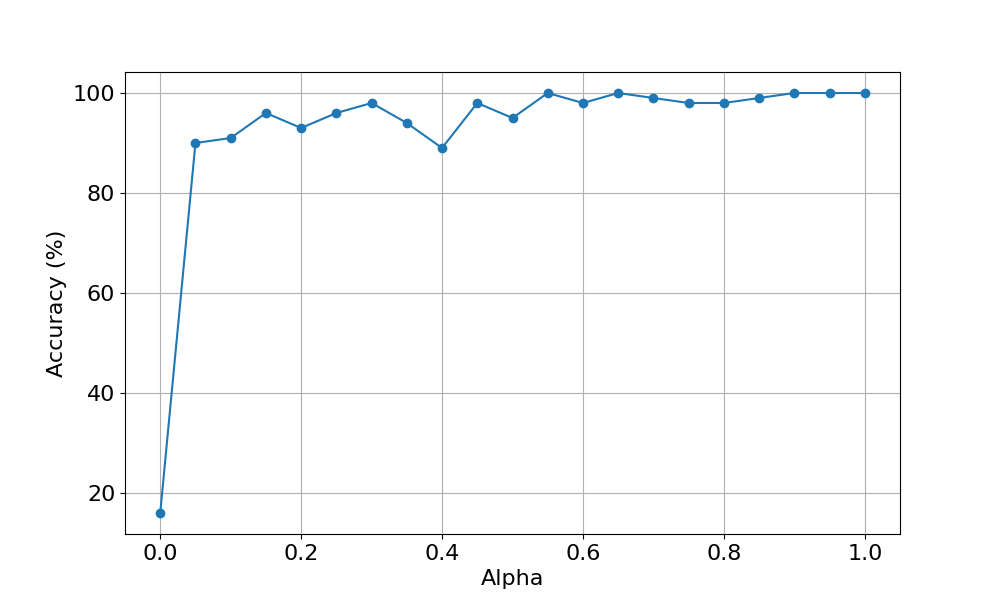}
		\label{fig:sub1}
	}
	\hfill
	\subfigure[2CH Synthetic Images]{
		\includegraphics[width=0.45\textwidth]{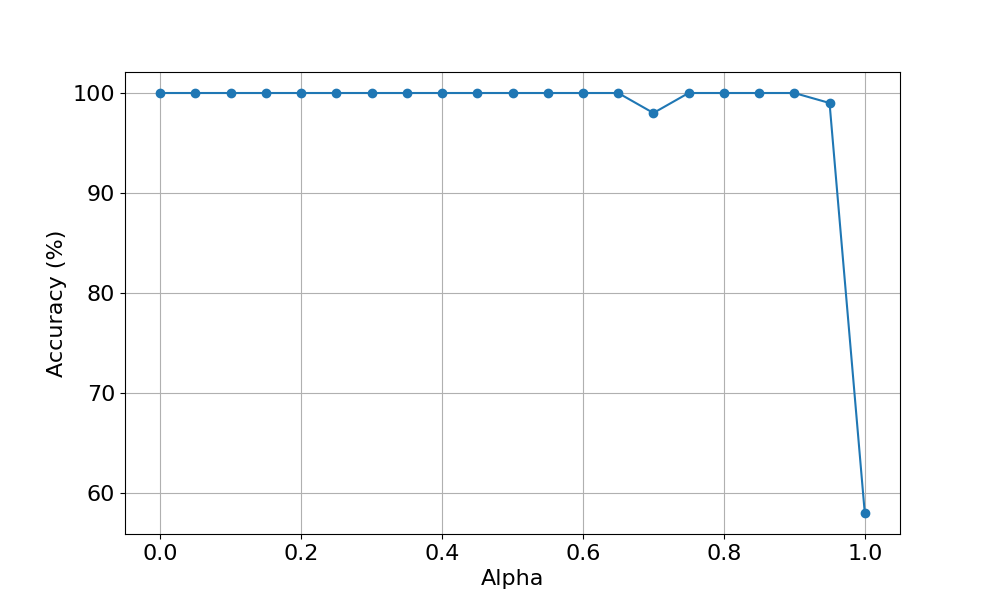}
		\label{fig:sub2}
	}
	
	\vspace{0.5cm}
	
	\subfigure[4CH Real Images]{
		\includegraphics[width=0.45\textwidth]{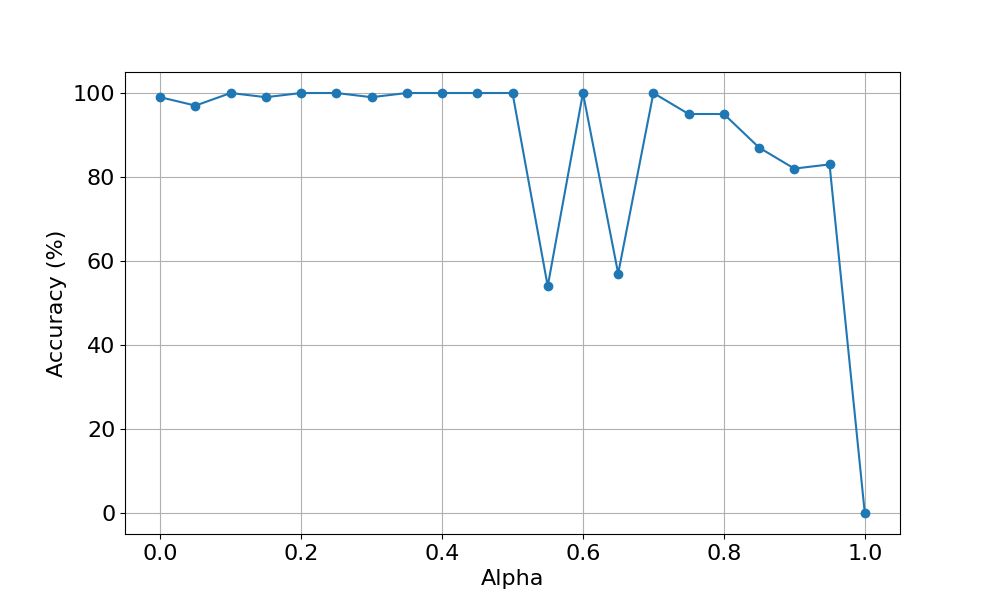}
		\label{fig:sub3}
	}
	\hfill
	\subfigure[4CH Synthetic Images]{
		\includegraphics[width=0.45\textwidth]{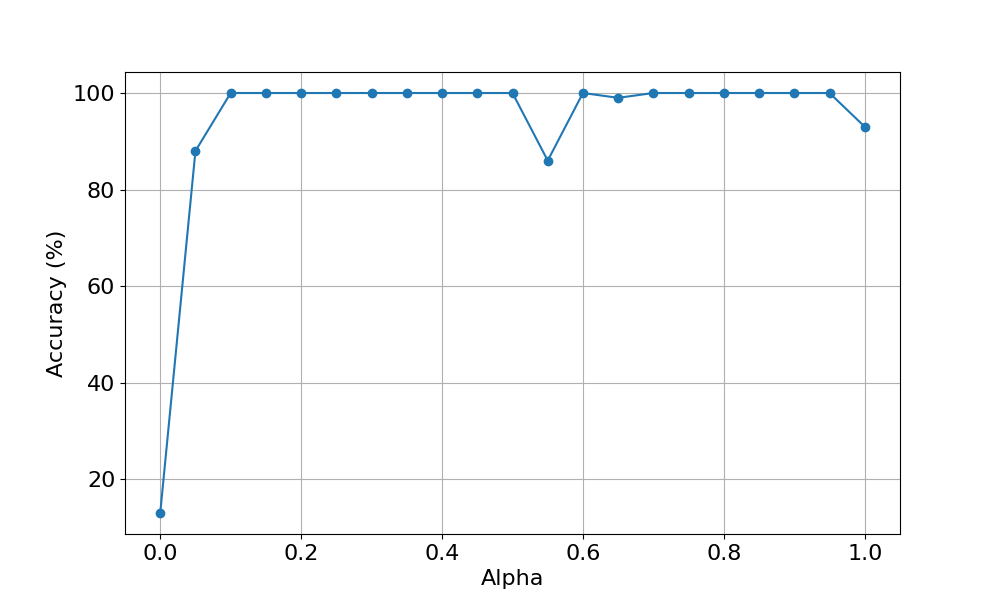}
		\label{fig:sub4}
	}
	
	\caption{Test Accuracy for the CAMUS Echocardiograms Classification with varying $\alpha$}
	\label{fig:camus_classification}
\end{figure}

\begin{figure}[ht]
	\centering
	
	\subfigure[Real Digit 2 Images]{
		\includegraphics[width=0.45\textwidth]{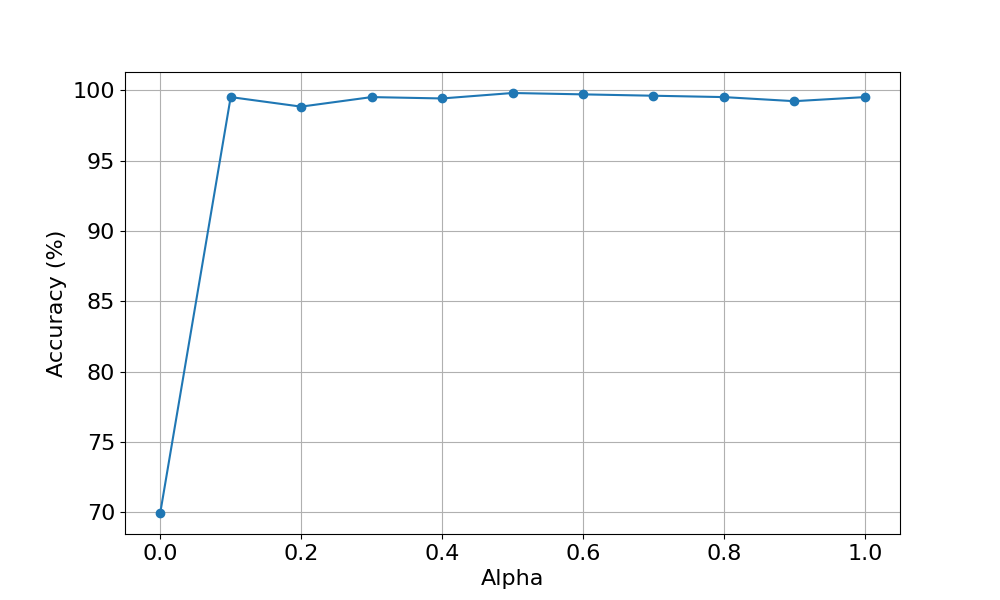}
		\label{fig:sub1}
	}
	\hfill
	\subfigure[Synthetic Digit 2 Images]{
		\includegraphics[width=0.45\textwidth]{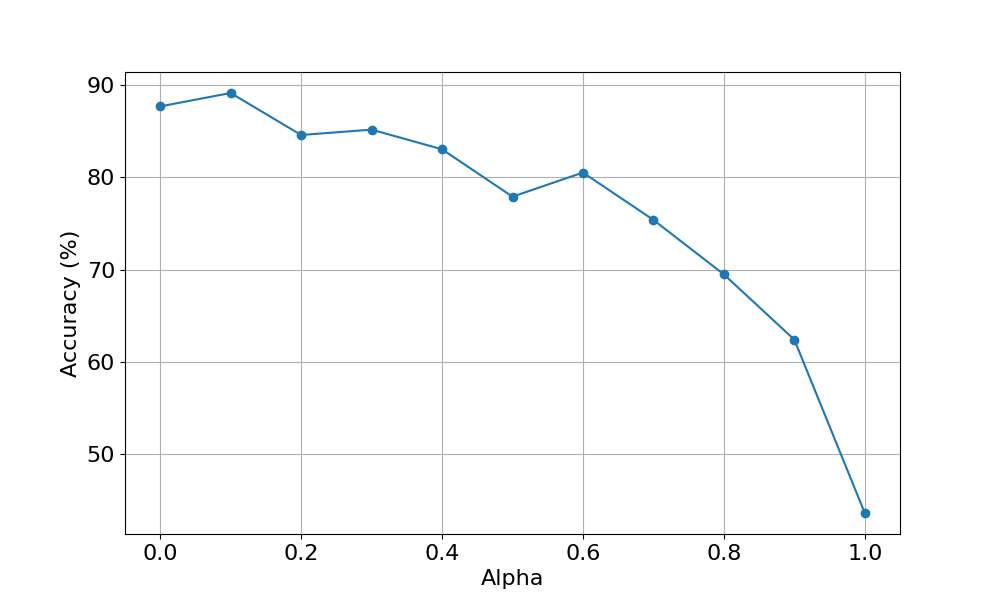}
		\label{fig:sub2}
	}
	
	\vspace{0.5cm}
	
	\subfigure[Real Digit 4 Images]{
		\includegraphics[width=0.45\textwidth]{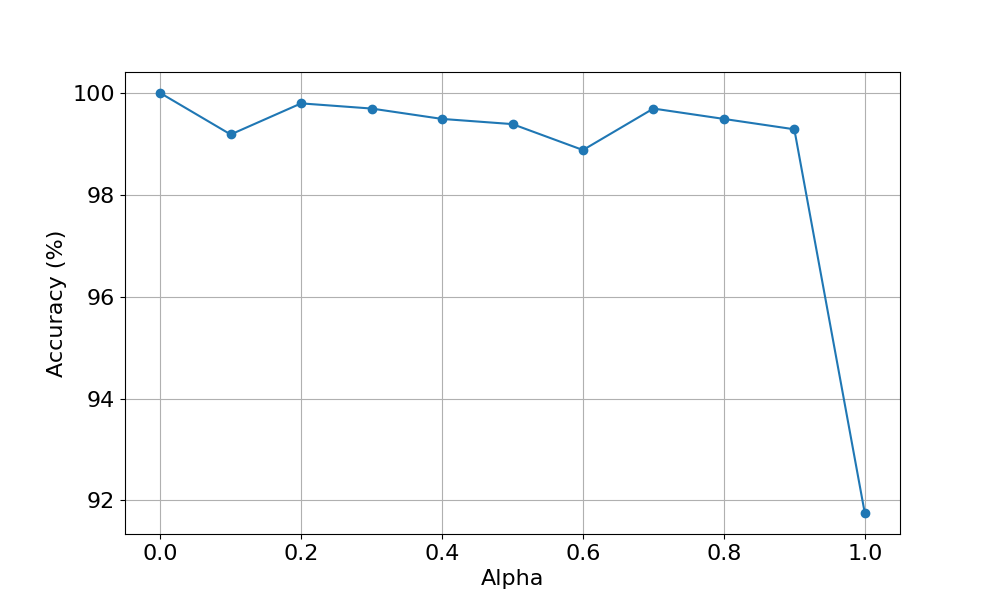}
		\label{fig:sub3}
	}
	\hfill
	\subfigure[Synthetic Digit 4 Images]{
		\includegraphics[width=0.45\textwidth]{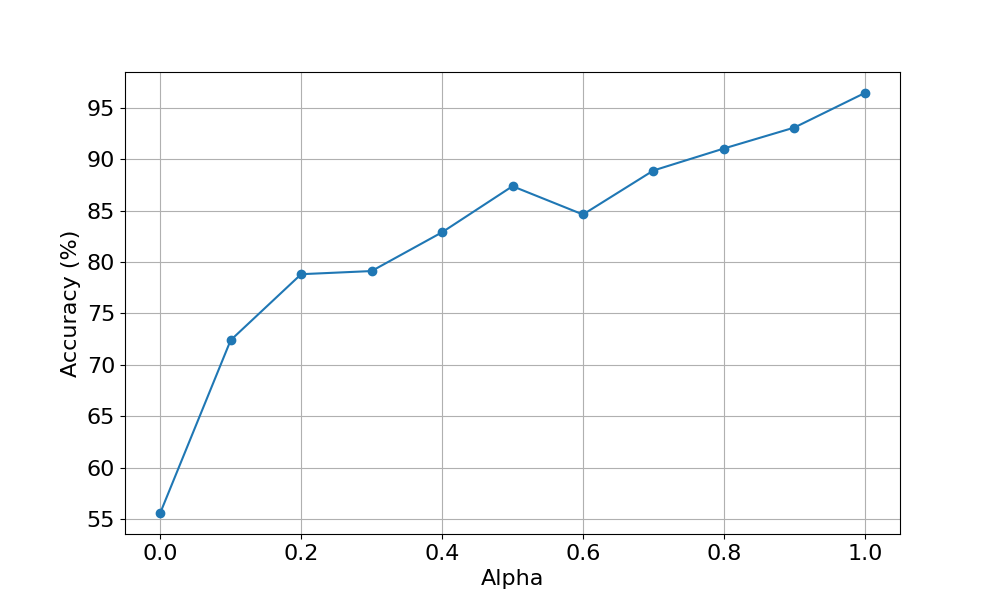}
		\label{fig:sub4}
	}
	
	\caption{Test Accuracy for the MNIST Digit Classification with varying $\alpha$}
	\label{fig:mnist_classification}
\end{figure}

Experimental results for the CAMUS and MNIST classification tasks are presented in Figures \ref{fig:camus_classification} and \ref{fig:mnist_classification}, respectively. The following observations can be noted:

\begin{enumerate}
	\item \textbf{Poor Performance in Extreme Cases of $\alpha$}: In the extreme case of $\alpha = 1$ (high positive correlation), the training dataset consists solely of real 2CH images and synthetic 4CH images. Under this condition, the model performs well on real 2CH images (Fig. \ref{fig:camus_classification}(a)) and synthetic 4CH images (Fig. \ref{fig:camus_classification}(d)). However, due to the simplicity bias, the model's performance on synthetic 2CH images (Fig. \ref{fig:camus_classification}(b)) and real 4CH images (Fig. \ref{fig:camus_classification}(c)) is nearly zero. 
	Similarly, in the extreme case of $\alpha = 0$ (high negative correlation), the model performs poorly on real 2CH images (Fig. \ref{fig:camus_classification}(a)) and synthetic 4CH images (Fig. \ref{fig:camus_classification}(d)). The MNIST results in Fig. \ref{fig:mnist_classification} show a similar trend.
	
	\item \textbf{Implication of Synthetic Simplicity in Medical vs./ General Domains}: We can observe that the drop in test accuracy for CAMUS in all four cases is significantly steeper and more sudden compared to MNIST, where the drop is more gradual.  We hypothesize that this variation between echocardiographic classification vs./ digit classification is due to the fact that the complexity of the former task is arguably more complex, which biases the network to rely more heavily on the simpler feature - source of data. This indicates that the impact of synthetic simplicity has more severe implications in the medical context compared to general contexts. The complexity of the feature set in medical image data likely contributes to this pronounced effect. This underscores the urgent need to address simplicity bias in the medical domain, where the stakes are considerably higher due to the critical nature of the applications. Ensuring that models generalize well and do not overly rely on simplistic and spurious features is crucial in medical imaging to avoid potentially harmful misclassification. 
	
	\item \textbf{Case of Balanced Augmentation}: When the correlation between the task labels and the data source is set to around 0.5, representing a balanced case, the test accuracy is quite consistent and is not significantly affected by the value of $\alpha$. This can be observed in both the CAMUS (Fig. \ref{fig:camus_classification}) and MNIST (Fig. \ref{fig:mnist_classification}) results. This observation suggests that in order to mitigate the simplicity bias  while using the synthetic data, we need to adopt a balanced approach when augmenting synthetic data with real data to achieve robust evaluation. A balanced approach ensures that the model does not overly rely on spurious features that are easy to learn but irrelevant to the actual task. By maintaining an intermediate level of correlation, the model is encouraged to utilize more complex and relevant features, leading to better generalization.
	
	\item \textbf{Persistence of Synthetic Simplicity Across Different Downstream Networks}:
	In our study, we employed a basic CNN architecture for digit classification and a more intricate ResNet model for echocardiogram view classification as the downstream network. Despite ResNet's increased complexity, we consistently observed synthetic simplicity across both tasks. This phenomenon persisted irrespective of the network's architecture and depth. Even with its deeper layers, ResNet tended to exploit simplistic and spurious features inherent in synthetic data, leading to diminished performance during testing. These findings highlight the pervasive nature of synthetic simplicity in downstream tasks and emphasize the need for cautious integration of synthetic data for augmentation purposes.
\end{enumerate}

\section{Conclusion}
\label{sec:conc}
We clearly demonstrate that downstream tasks can leverage the binary feature of the dataset source (real v/s synthetic) for tasks such as classification. Therefore, it is crucial for practitioners to use synthetic data carefully during model development. This caution helps prevent the inclusion of spurious features and the generation of artificially high accuracy values on a test set derived from the same distribution. We hope that this work will stimulate more interest in understanding the risks associated with using synthetic data without domain expertise and encourage further research.

%
%
%
 \bibliographystyle{splncs04}
 \bibliography{mybibliography}
\end{document}